\PassOptionsToPackage{table}{xcolor} 
\documentclass[sigconf,nonacm]{aamas}

\usepackage{balance} 
\usepackage{multirow} 
\usepackage{array}   
\newcolumntype{L}[1]{>{\raggedright\arraybackslash}p{#1}}

\newcommand{\symok}{\raisebox{-0.16em}{\includegraphics[page=1,height=0.85em]{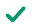}}}
\newcommand{\symx}{\raisebox{-0.16em}{\includegraphics[page=2,height=0.85em]{figs/sym-marks}}}
\newcommand{\symprev}{\raisebox{-0.16em}{\includegraphics[page=3,height=0.85em]{figs/sym-marks}}}
\newcommand{\symcond}{\raisebox{-0.16em}{\includegraphics[page=4,height=0.85em]{figs/sym-marks}}}

\title{guardscope: Measuring Which Actions Runtime Controls Check in LLM Agent Systems}

\acmConference[Preprint]{Preprint manuscript}{November 2026}{}{}

\author{Kentaroh Toyoda}
\affiliation{
  \institution{Vulcan Research, AIFT, Singapore}
  \country{}
}
\affiliation{
  \institution{Keio University Global Research Institute, Japan}
  \country{}
}
\email{kentaroh.toyoda@aift.io}

\author{Alex Leung}
\affiliation{
  \institution{Vulcan Research, AIFT, Singapore}
  \country{}
}
\email{alex.leung@aift.io}

\author{Bhuvaneswari U}
\affiliation{
  \institution{Standard Chartered, India}
  \country{}
}
\email{bhuvaneswari.1@sc.com}

\begin{teaserfigure}
  \begin{center}
    \includegraphics[width=0.3\columnwidth]{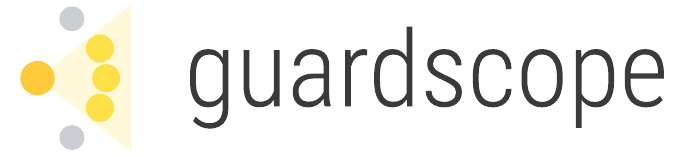}
    \Description{The guardscope logo: the guardscope wordmark with a yellow logo mark.}
  \end{center}
\end{teaserfigure}

\begin{abstract}
  Large-language model (LLM) agents execute actions on external systems through tool calls, and enterprises deploy runtime controls, such as agent gateways, to check this behavior in real time. However, a control can only evaluate the actions that cross the point where it is placed, and an agent can act through more paths than the tool-call channel; no existing work measures this coverage against kernel-level ground truth. In this paper, we present guardscope, a framework that makes this coverage measurable. guardscope works in two stages: a static stage that predicts which of an agent's 15 action paths a control can check from its source, and an empirical stage that runs a fixed workload on a deployment that realizes one of eight topology designs and classifies every executed action as checked, bypassed, or prevented by reconciling the control's decision records against kernel-level ground truth. guardscope makes a control's coverage claim verifiable: a runtime control developer can publish which paths its product inspects under a kernel-verified protocol. We conducted experiments with two agent harnesses and real runtime controls, including Microsoft's agent governance toolkit and agentgateway. guardscope is available at \url{https://github.com/VulcanLab/guardscope}.
\end{abstract}



\begin{document}

\maketitle

\section{Introduction}

Large-language model (LLM) agents execute actions on external systems through tool calls, and enterprises deploy them for work that touches internal data and services. An enterprise AI system must follow the enterprise's security and compliance policies, the same as any other enterprise system. Therefore, a growing ecosystem of runtime controls checks agent behavior in real time.

These controls occupy different positions in the system. Agent gateways, such as agentgateway~\cite{agentgateway}, Envoy AI Gateway~\cite{envoyaigateway}, and Microsoft's Agent Governance Toolkit (AGT)~\cite{agt}, interpose a check on the traffic between the harness and its tool servers. Guardrail engines, such as NeMo Guardrails~\cite{nemo} and LlamaFirewall~\cite{llamafirewall}, screen model inputs and outputs against content and safety policies. Monitoring tools, such as AgentSight~\cite{agentsight} and Tetragon~\cite{tetragon}, observe agent behavior from outside the agent process. Policy engines, such as Progent~\cite{progent} and ActPlane~\cite{actplane}, decide whether a concrete action may execute.

However, a runtime control can only evaluate the actions that cross the point where it is placed. A gateway in front of the harness sees tool requests with full semantic detail, but the same harness can spawn tool servers as local child processes, and agent code can open network connections directly, so actions can take effect without crossing the gateway. Kernel-level monitoring with extended Berkeley Packet Filter (eBPF) observes nearly every action of the agent process tree, yet its records carry no payload content and mix in processes unrelated to the agent. Consequently, a control's placement determines both which actions it can check and what information it can inspect, but no existing work measures this placement-driven coverage of deployed controls against kernel-level ground truth.

In this paper, we make this relationship measurable. We enumerate 15 action paths that span the routes a model's decision can travel from the harness runtime to its effect, and we identify eight topology designs that span the enforcement positions used in practice, from a gateway in front of the harness to enforcement inside the harness process and in the operating-system kernel. For each pair of path and design, we derive whether the design checks the path. We then present guardscope\footnote{\url{https://github.com/VulcanLab/guardscope}}, a framework that turns this analysis into a repeatable measurement. guardscope works in two stages. The static stage analyzes a runtime control's source and predicts which of the 15 paths the control can check. The empirical stage runs a fixed workload of 15 action instances on a deployment that realizes one design, records what the control reports, and verifies the record against what the host kernel observed. The comparison classifies every action instance as checked, bypassed, or prevented.

We evaluate guardscope across the eight designs with Pi and OpenClaw, and with real runtime controls: Microsoft's agent governance toolkit as the gateway control and ActPlane~\cite{actplane} as the kernel-level control, with a Tracee eBPF sensor providing the kernel-level ground truth. The runs confirm the derived outcomes for the exercised action classes. We make the following contributions:
\begin{itemize}
  \item an enumeration of the 15 action paths and the eight topology designs, with the derived matrix of which paths each design checks and the trade-off between coverage and inspectable features;
  \item guardscope, a two-stage framework that predicts a runtime control's checkable paths by static analysis and measures instance-level coverage against kernel-level ground truth;
  \item an evaluation that realizes the eight designs with real runtime controls and two harnesses, confirming the derived outcomes for the exercised action classes and showing outcome stability across the harness.
\end{itemize}

The remainder of this paper is organized as follows. Section~\ref{sec:related} discusses related work. Section~\ref{sec:paths} enumerates the action paths and the topology designs. Section~\ref{sec:guardscope} presents guardscope. Section~\ref{sec:evaluation} reports the evaluation. Section~\ref{sec:limitations} states the limitations, and Section~\ref{sec:conclusion} concludes.

\section{Related Work}
\label{sec:related}

We classify prior work into five categories: (1) observing behavior with monitoring, (2) checking tool traffic and enforcing policy, (3) screening content with guardrails, (4) restricting actions with access control, and (5) evaluating or measuring enforcement. 

\subsection{Behavior Monitoring}

The first category is behavior monitoring, in which runtime controls observe agent behavior. AgentOps~\cite{agentops} defines the observability data to collect from LLM-agent runs. 
ClawSentry~\cite{clawsentry} delivers monitoring through a gateway.
Watson~\cite{watson} traces the reasoning of LLM-powered agents. System-level approaches observe from outside the application process. 
AgentSight~\cite{agentsight} intercepts LLM traffic and kernel events with eBPF and correlates the two streams across process boundaries. eBPF-based audit tools such as Tetragon~\cite{tetragon} and Falco~\cite{falco} provide the same kernel-wide record of process activity for general workloads. 

\subsection{Agent Gateways and Policy Checks}

The second category is agent gateways that monitor interactions between the harness and its tools by intercepting them at a network level. Envoy AI Gateway~\cite{envoyaigateway} and Apache APISIX expose extensible filter chains to which policy modules attach. Agentgateway~\cite{agentgateway}, LiteLLM Proxy~\cite{litellm}, Bifrost~\cite{bifrost}, AISIX~\cite{aisix}, Portkey~\cite{portkey}, Pomerium~\cite{pomerium}, Octelium~\cite{octelium}, Plano~\cite{plano}, and MCP Context Forge~\cite{contextforge} apply the same modular structure to LLM-specific traffic. Microsoft's Agent Governance Toolkit (AGT) adds governance modules to the gateway function, including deterministic policy decisions on every tool call, tool-definition scanning, and checks for tool servers that connect over the Model Context Protocol (MCP). Cloud control planes enforce equivalent tool-call policy inside the provider's service boundary rather than on a proxy deployed between the harness and the tool server. For instance, Microsoft Foundry~\cite{msfoundry} applies runtime guardrails at the tool-call and tool-response steps of its agent service, and Agent 365~\cite{agent365} routes registered tool servers through its tooling gateway, which enforces policy at invocation time. The policies these components evaluate are written in the policy-as-code ecosystem, including Rego with Open Policy Agent~\cite{opa}, Cedar~\cite{cedar}, and CEL~\cite{cel}. 

\subsection{Guardrails}

The third category is guardrails that screen model inputs and outputs against content and safety policies.\footnote{We cannot summarize recent guardrail tools and papers here due to space. The interested reader is referred to recent survey papers, such as \cite{buildingguardrails, fullstacksafety, guardrailssurvey}.} Programmable guardrail toolkits define check sequences around the model call. For instance, NeMo Guardrails~\cite{nemo} provides input moderation, output moderation, and execution checks specified in a dedicated language. Security-focused frameworks compose several detectors: LlamaFirewall~\cite{llamafirewall} combines a jailbreak detector, an alignment check against the agent's stated goal, and code scanning for insecure agent outputs. Model-based classifiers perform the screening with a dedicated model, such as Llama Guard~\cite{llamaguard}, and data-screening components such as Presidio~\cite{presidio} detect and redact sensitive content in tool payloads. Agent-based guardrails reason over agent trajectories. GuardAgent~\cite{guardagent} maps a safety guard request into guardrail code that checks target-agent actions, and ShieldAgent~\cite{shieldagent} extracts verifiable safety rules from policy documents and evaluates agent action sequences against them. 

\subsection{Access Control}

The fourth category is access-control systems that decide whether a concrete action may execute. Most systems in this category embed the decision procedure in the agent's own process, and they fall into three families: (1) classic access-control systems, (2) capability-based systems and (3) policy-language systems. Classic access-control systems apply access-control rules directly to agents. Ganie~\cite{rbacagent} integrates role-based access control into agent deployments, and AgentGuard~\cite{agentguard} applies attribute-based access control, checking every tool call against declarative policies. 

Capability-based systems derive the decision from information flow. For instance, CaMeL~\cite{camel} extracts capabilities from the trusted user query and enforces control-flow and data-flow policies over them. RTBAS~\cite{rtbas} applies information-flow control to tool-based agents with dependency screeners. APPA~\cite{appa} extends information-flow control with recovery mechanisms. Leong~\cite{leong} implements an external monitor with capability-gated execution. Harness-native enforcement places the monitoring point in the harness configuration. SHarD~\cite{shard} shows that OS sandboxing, skill scanning, and tool restriction can be distributed through a harness. SkillGuard~\cite{skillguard} derives capability confinement from reachability analysis. SEAgent~\cite{seagent} applies mandatory access control to privilege-escalation scenarios in multi-agent systems. 

Policy-language systems instead decide from declarative rules over the action's features. For instance, Progent~\cite{progent} enforces least privilege through symbolic rules over tool names and arguments, with monotonic policy narrowing during execution. AgentFlow~\cite{agentflow} defines a flow-centric policy language with stateful taint semantics and a verifier based on satisfiability modulo theories (SMT), which encodes policy constraints as logical formulas and decides them with a solver. FORGE~\cite{forge} specifies policies in Datalog, a declarative logic-programming language, over abstract predicates that describe the execution context and constrains the predicate source with an assume/guarantee contract. AgentSpec~\cite{agentspec} specifies runtime constraints as rules over triggers, predicates, and enforcement mechanisms, and evaluates them in the agent's decision pipeline at action-execution time. Kernel-level enforcement moves it into the operating system. ActPlane~\cite{actplane} enforces agent policies with eBPF and reports that tool-call interception misses system actions on indirect execution paths. 

\subsection{Evaluation and Benchmarks}

The fifth category is evaluation and benchmarks. Agent benchmarks such as AgentDojo~\cite{agentdojo} execute attack and defense implementations against shared tasks to measure prompt-injection attacks and defenses. The adaptive evaluation study~\cite{adaptiveeval} is the only published adaptive test of a policy-based monitor. It tested Progent on AgentDojo with a single attack template, and the defense held. Several papers report individual attacks in which an agent acts within its authority and the action is harmful. Causality laundering~\cite{causality} extracts information from denial outcomes, the visual confused deputy~\cite{visualdeputy} exploits the divergence between what a computer-using agent perceives and what it acts on, and cross-app context poisoning~\cite{confusedchatgpt} injects content through first-party interfaces of a shared agent context. A separate line measures policy-as-code as an artifact: adoption across 399 open-source repositories~\cite{foalem}, security misconfigurations in Kubernetes manifests~\cite{rahman}, and inconsistency among Helm chart analyzers~\cite{minna}. The design-pattern catalog~\cite{designpatterns} assesses the prompt-injection resistance of proposed enforcement architectures, including gateways and dual-model configurations.

\section{Action Paths and Topology Designs}
\label{sec:paths}

The related work shows that runtime controls occupy several enforcement positions, from gateways in front of the harness to the operating-system kernel, but it does not state which of an agent's actions each position can actually check. In this section, we enumerate the action paths an agent can take and the topology designs used in practice, and we derive the coverage outcome for each pair of path and design. guardscope (Section~\ref{sec:guardscope}) then turns this analysis into predictions and measurements.

\subsection{Action Paths}

\begin{table*}[t]
  \centering
  \caption{Paths that agent harnesses can take for execution.}
  \label{tab:paths}
  \includegraphics[width=\textwidth]{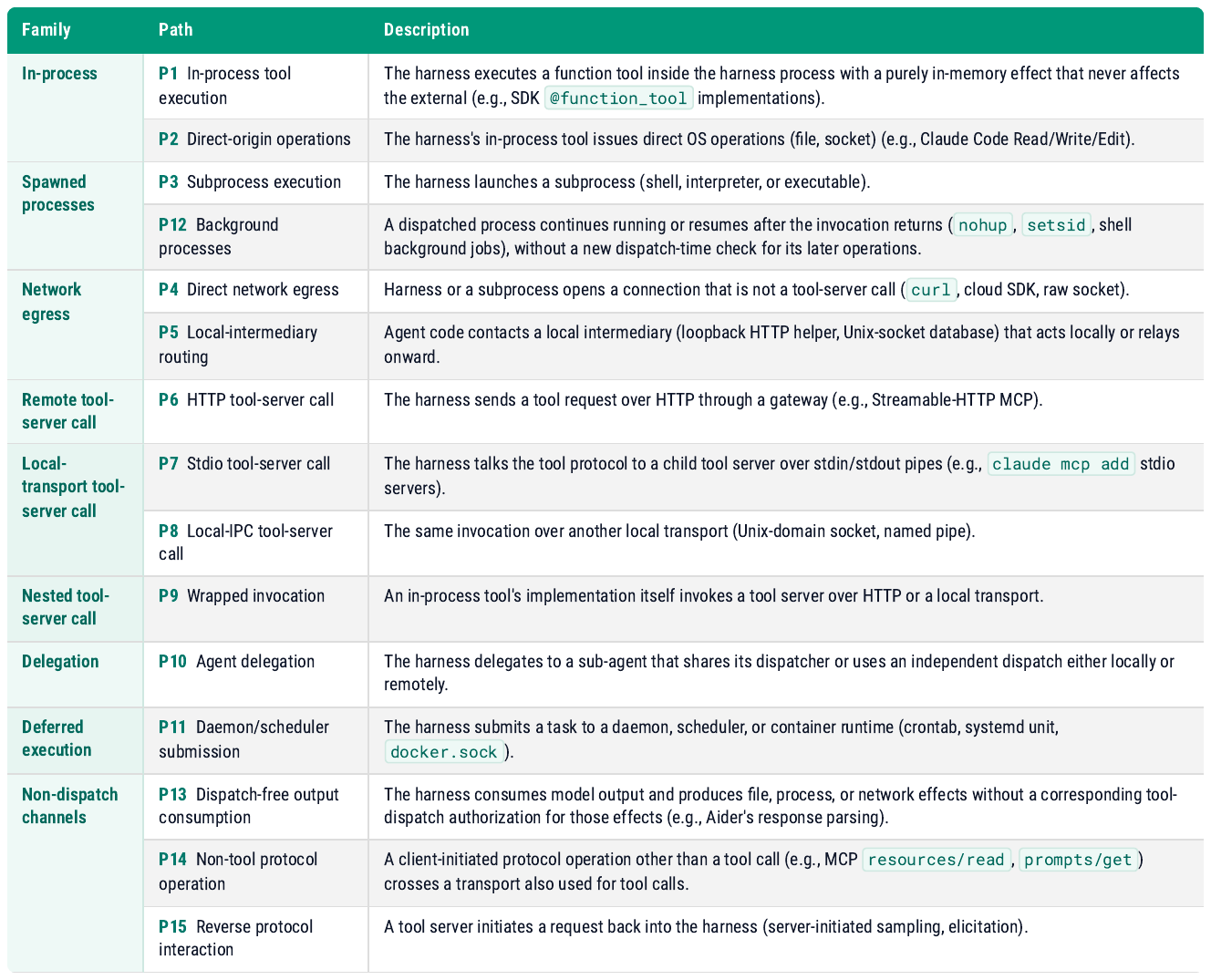}
\end{table*}

Table~\ref{tab:paths} enumerates the paths that an agent harness can take for execution. We derived this basis bottom-up, from the routes by which executed actions reach their effects in real harnesses, and we found a total of 15 potential paths with nine family classes. We developed the basis in an iterative discussion between us and two AI research models, Z.ai's GLM-5.3 and OpenAI's ChatGPT 6.1-Sol: each side independently proposed a candidate path list, cross-reviewed the others' proposals, and converged on the final basis.

Most of the paths in Table~\ref{tab:paths} name a transport directly, so we supplement only the splits a reader might contest, and name what separates them. In-process tool execution (P1) and direct-origin operations (P2) share the same code location, but P1 is a dispatched tool whose effect stays in memory, whereas P2 is an OS operation that the harness process issues directly, with no dispatch for a harness-native design to inspect; the effect, not the location, separates them. The three process paths form the closest group: subprocess execution (P3) and background processes (P12) share a spawned child, but a P3 child is joined within the invocation, so its operations fall under the invocation's check, whereas a P12 child outlives the invocation and its later operations receive no dispatch-time check; lifetime separates them. Daemon submission (P11) also produces later execution, but the executor sits outside the harness's process tree and the submission is a record, so attribution separates it from P12. Dispatch-free output consumption (P13) and direct-origin operations (P2) both produce effects without a dispatch, but a tool call exists in P2 and none exists in P13; whether a tool call exists separates them.

\subsection{Topology Designs}

\begin{figure*}[t]
  \centering
  \includegraphics[width=\textwidth]{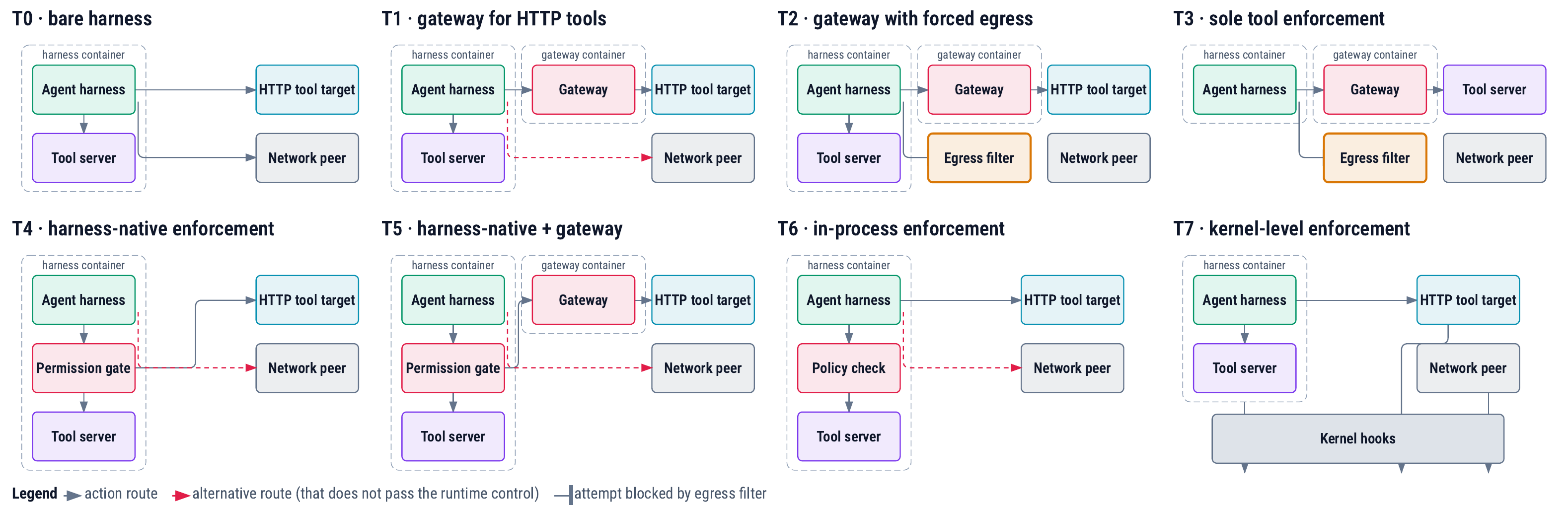} 
  \caption{System architecture of topology designs.}
  \label{fig:topologies}
\end{figure*}

\begin{table*}[t]
  \centering
  \caption{Topology designs and inspectable features.}
  \label{tab:topologies}
  \includegraphics[width=\textwidth]{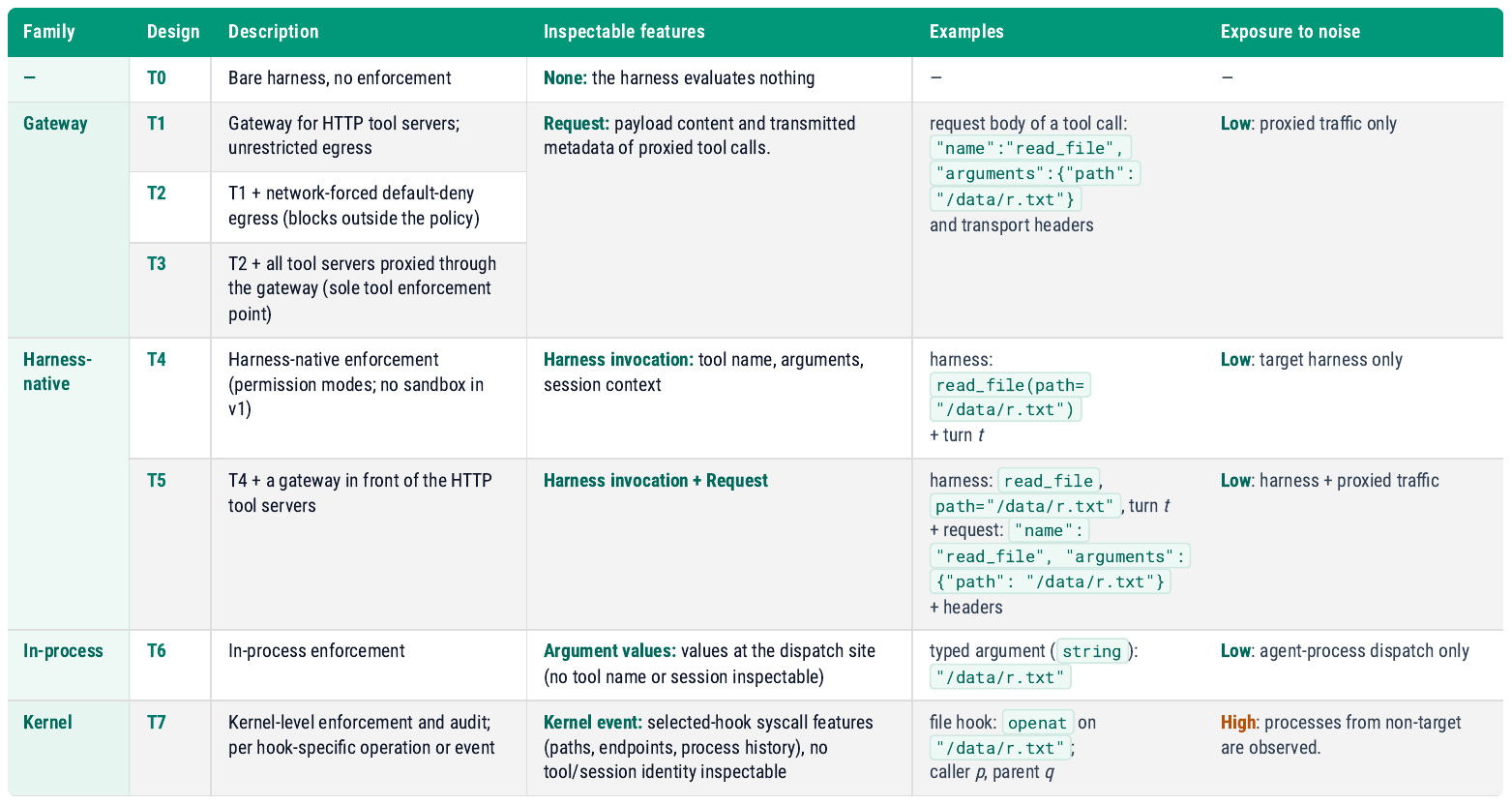}
\end{table*}

We identified the topology designs from the enforcement positions used in practice: across the deployed systems surveyed in Section~\ref{sec:related}, we collected the placements at which a runtime control can sit, from no control at all (T0), through gateways in front of the harness, the harness's own enforcement mechanisms, and in-process policy engines, to the operating-system kernel (T7), and we generalized them into the eight designs shown in Figure~\ref{fig:topologies}. Every design generalizes the configuration of at least one deployed system (Table~\ref{tab:tiermap}), which keeps the design space tied to practice rather than to an abstract taxonomy. The designs went through the same AI-assisted review as the path basis (Section~\ref{subsec:ai-usage}).

Two groups of designs share a placement, so we state what separates them. The gateway designs (T1--T3) share the placement in front of the harness, and what separates them is what the deployment additionally routes through or blocks at the gateway: T2 adds a forced-egress environment that prevents direct network egress, and T3 additionally routes the local transports through the gateway. A design is therefore a placement plus the deployment's routing and blocking environment, not a placement alone. The in-harness designs (T4--T6) share the placement inside the harness, and what separates them is the enforcement mechanism and what it can inspect: T4 is the harness's own permission chain, T5 composes that chain with a gateway in front, and T6 is a declarative policy engine evaluated at the harness's dispatch hook, which can inspect tool arguments that a permission mode cannot. Table~\ref{tab:topologies} lists the designs with the features that the runtime controls can inspect and their examples.

\subsection{Derived Coverage Outcomes}

\begin{figure}[t]
  \centering
  \includegraphics[width=.9\columnwidth]{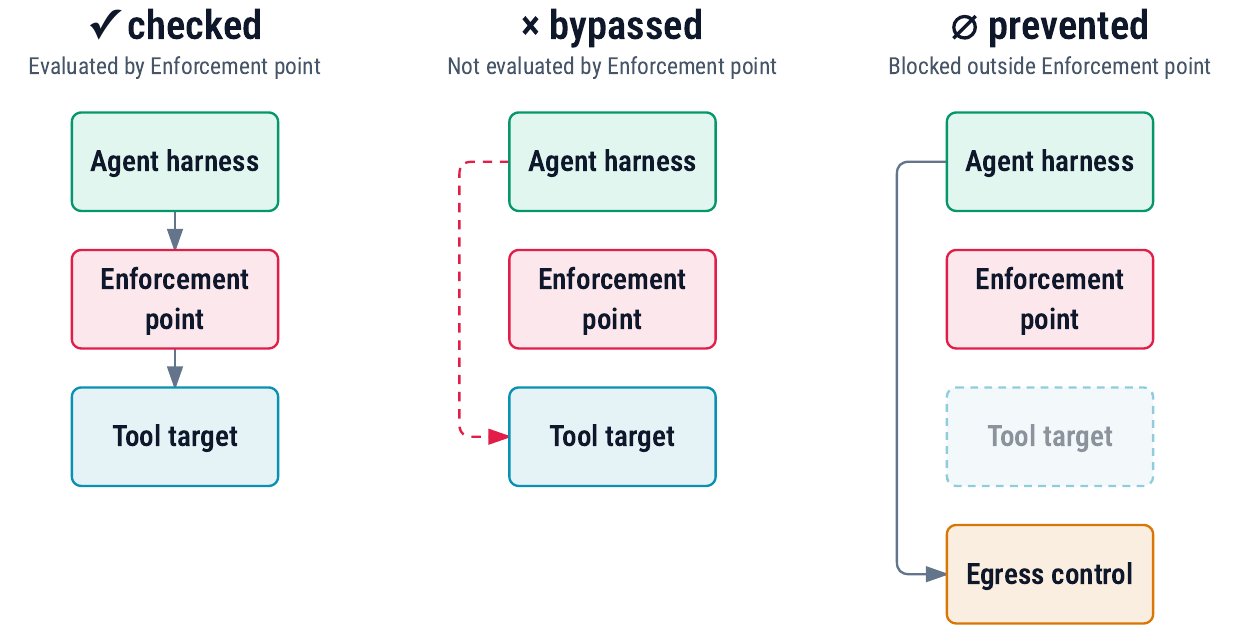}
  \caption{The three instance-level coverage outcomes.}
  \label{fig:three-coverage-types}
\end{figure}

Against each pair of path and topology design, we systematically derive whether the design checks the path, and whether some mechanism outside the policy prevents it. Here, Figure~\ref{fig:three-coverage-types} shows the definition of ``checked (\symok),'' ``bypassed (\symx),'' or ``prevented (\symprev).'' An action path is checked if its channel crosses an enforcement point. 
In contrast, an action is bypassed if its channel crosses no enforcement point and no blocking mechanism. 
Lastly, an action is prevented if its channel is blocked by a mechanism that the topology places outside the policy, such as a network egress filter.  An exception is ``conditional (\symcond),'' where the results depend on the actual systems or deployments. Table~\ref{tab:pathmatrix} shows the derived per-path expected outcomes under the eight topology designs. We see from Table~\ref{tab:pathmatrix} that the coverage of a design is determined by the paths that cross its enforcement point. The gateway designs check only the tool-server calls that traverse the gateway (T1: P6, T3: P6--P8), and the forced egress of T2 and T3 additionally prevents direct network egress (P4). The harness-native and in-process designs (T4--T6) check the actions dispatched through the harness dispatcher, but not the actions that originate outside the dispatch, such as direct-origin operations (P2) or processes that outlive the invocation (P12). In contrast, the kernel-level design (T7) monitors all the paths except the harness's in-process actions (P1), which the harness-native controls (T4--T6) capture. As a result, five paths (P2, P5, P11, P12, and P15) bypass every design below the kernel. The T7 row states the bound implied by the kernel position: the coverage a deployed kernel-level control actually provides depends on the hooks it deploys, and we quantify this gap in Section~\ref{subsec:results}. 

Table~\ref{tab:pathmatrix} also marks several cells as conditional (\symcond). A wrapped invocation (P9) is checked under the gateway designs only when the wrapper itself sends the inner call through the gateway, and a non-tool protocol operation (P14) is checked only when the gateway terminates the protocol and evaluates the operation. Agent delegation (P10) is checked when the sub-agent shares the dispatcher or inherits the enforcement configuration, and dispatch-free output consumption (P13) is caught by the forced egress only when its downstream effect is a network connection. Under T7, a daemon submission (P11) is observed only when the eventual execution crosses a monitored hook. These outcomes depend on the deployment wiring rather than on the design. The reference workload realizes four of the 15 paths (P2, P4, P6, and P7), so the conditional cells rest on the design analysis alone.

\begin{table}[t]
  \centering
  \caption{Derived per-path expected outcomes under the eight designs.}
  \label{tab:pathmatrix}
  \includegraphics[width=\columnwidth]{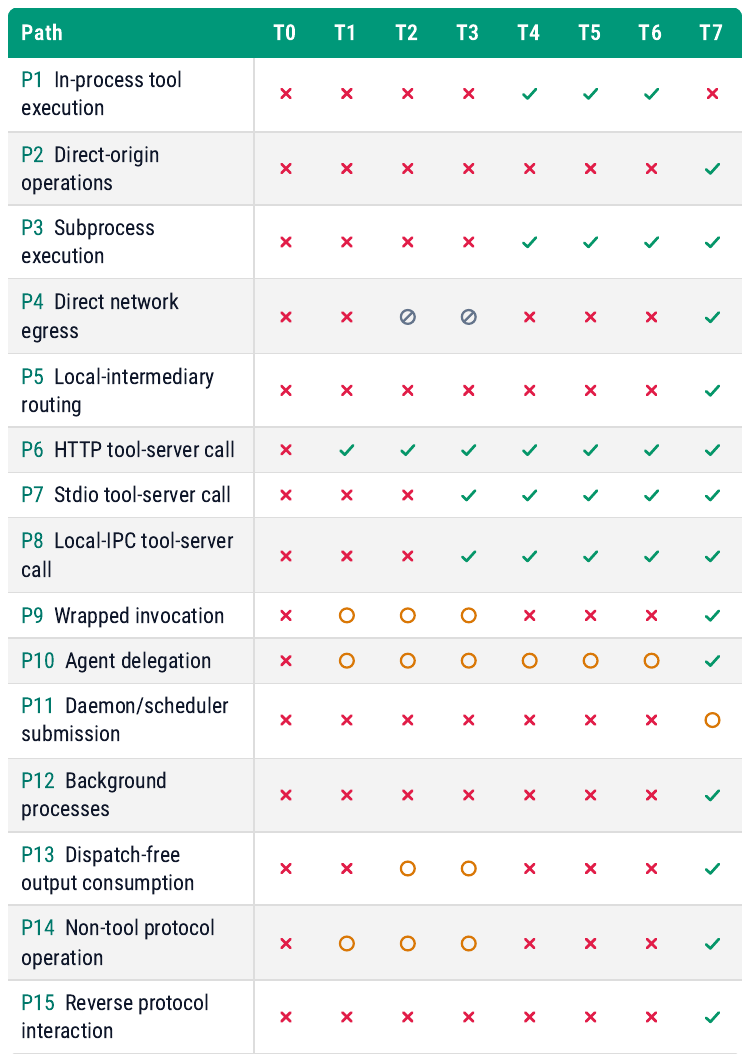}
\end{table}

\subsection{Trade-offs between Coverage and Inspectable Features}
\label{subsec:tradeoffs}

From Tables~\ref{tab:pathmatrix} and~\ref{tab:topologies}, we can see that there are trade-offs between coverage, inspectable features, and exposure to noise. The gateway designs inspect the payload content and metadata of the proxied requests, but they see few paths (T1 checks one). The harness-native design (T4) checks five paths and inspects the tool name, arguments, and session context. The kernel-level design (T7) checks the most paths (13 of the 15 unconditionally; the in-process tool execution (P1) is the only path outside the kernel position, and the daemon submission (P11) is conditional on the deployed hooks), yet it inspects only syscall-level features, such as file paths and endpoints, without tool or session identity, and it observes processes unrelated to the agent. In addition, T7 has three drawbacks: its inspectable features are the least semantic (e.g., \texttt{openat} and its file path attribute), they are noisy (i.e., T7 observes information from non-target processes), and the deployment cost is higher, as it must be deployed in the same system as the target agent.

\section{guardscope}
\label{sec:guardscope}

The path and design analysis of Section~\ref{sec:paths} states what a deployment should cover on paper. To verify what a deployed control actually covers, we built guardscope, a framework that predicts a runtime control's checkable paths from its source and measures, against kernel-level ground truth, which executed actions the control evaluates.
\subsection{Scope and Assumptions}
\label{subsec:assumptions}

Throughout the paper, we assume generic agent harnesses, such as Claude Code, Gemini CLI, Codex, DeepSeek Harness, Pi, LangChain, and OpenClaw. To monitor them, we assume generic runtime controls as we have reviewed in Section~\ref{sec:related}.

We assume the generic single-agent scenario where an agent often works as a coordinator. Such an agent is composed of a model and a harness with tool-use capabilities. The reason for the single-agent assumption is that, technically, our analysis can be applied to multi-agent scenarios where multiple agents coordinate together to accomplish given tasks, as a runtime control tool can be applied to every single agent.\footnote{Note that we do not rule out the case where the agent spawns sub-agents via its tool-use capability.} The interaction between agents is established via network interfaces, and our path analysis later on covers such interfaces, too.

\subsection{Static Analysis}
\label{subsec:static}

The static stage analyzes a runtime control's source and predicts, for each of the 15 paths, whether the control can check the path. The analysis first classifies the control: the deployment forms it can take (a container gateway, an attached process, an embedded policy engine, a kernel control, or a log ingest), whether it enforces or only observes, the action classes it can parse, the record format it emits, and the designs of Table~\ref{tab:topologies} it can occupy. For each path, the analysis then works through a fixed checklist: what the control must intercept to check the path, which code evidence counts, and a verdict per path, checkable, not-checkable, conditional, or uncertain, with file-level evidence attached to every verdict. The analysis cross-checks the verdicts against the per-design dispositions that the path basis of Section~\ref{sec:paths} predicts. The output is executable: the capability manifest that declares what the control can do, and the binding overlays that deploy it into each design's fixture. The framework validates both before any run.

\begin{table*}[t]
  \centering
  \caption{Classification of deployed systems onto the designs of Table~\ref{tab:topologies}. }
  \label{tab:tiermap}
  \footnotesize
  \includegraphics[width=\textwidth]{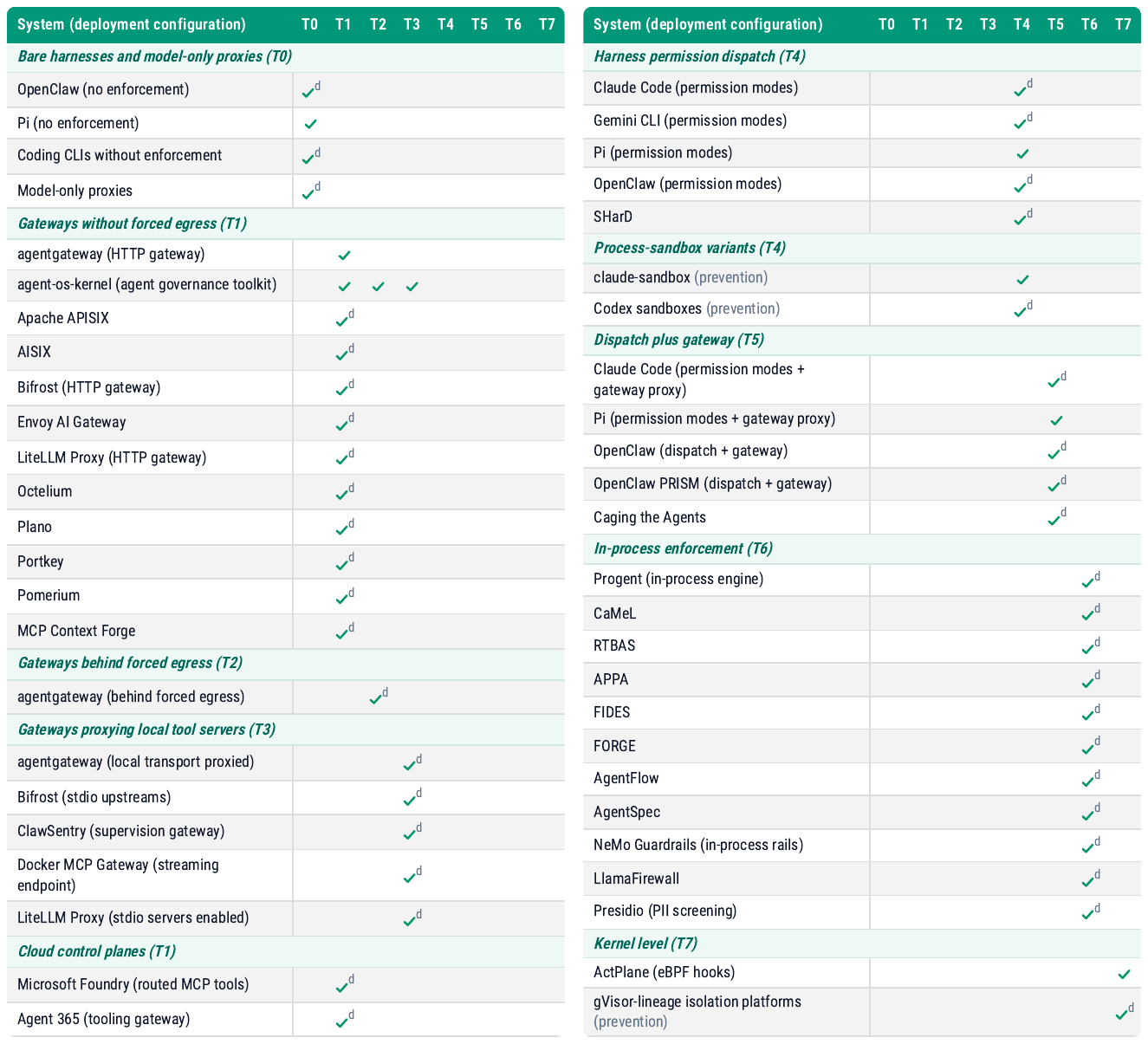}%
\end{table*}

We applied the static analysis to the deployed systems reviewed in Section~\ref{sec:related}; Table~\ref{tab:tiermap} shows the classification. In Table \ref{tab:tiermap}, we define two symbols, \symok\: classification is verified by our actual runs and \symok$^{d}$: classification is derived from documentation only. The measured rows correspond to the deployments we realize in the evaluation (Section~\ref{sec:evaluation}); the remaining rows are derived from the controls' source and documentation.

We can see from Table~\ref{tab:tiermap} that most of the deployed solutions take the designs with narrower coverage. The gateway deployments (T1--T3) form the largest group, whereas the kernel-level design (T7) has only one entry. This is understandable from the trade-off analysis in Section~\ref{subsec:tradeoffs}, too. Note that the same system appears under multiple designs, such as agent-os-kernel (T1, T2, and T3) and Claude Code (T4 and T5), depending on its deployment configuration. 

\subsection{Empirical Measurement}
\label{subsec:empirical}

The empirical stage runs the reference workload on a deployment that realizes one design and measures which executed actions the deployed control evaluates. The workload consists of 15 action instances that realize the four exercised paths of Table~\ref{tab:paths} (the HTTP and stdio tool calls, the direct-origin operations, and the direct network egress). The runner drives a real agent harness by deterministic replay injection, so no LLM runs in the loop and every run executes the same actions. The runner enforces a fixed lifecycle: it provisions and configures the topology, proves that each enforcement point is ready, opens the measurement window, injects the workload, closes the window, and collects the records. The readiness gate is strict: the runner issues one action that the loaded policy must deny, and the run proceeds only when the enforcement point refuses it and its deny record appears in the collected stream. A started container is not a ready one, and opening the window before enforcement works would silently produce false bypassed results. Only the runner controls the window, never a measured component.

Each component emits its records in a common schema: an enforcement point emits one decision record per decision, and the harness emits dispatch records that identify each attempt. Correlation ids travel on the protocol, in the MCP \texttt{\_meta} block or in HTTP headers, and a control echoes them from the protocol rather than learning them from configuration, so a fabricated id is detectable. For the escape classes, whose traffic crosses no inspection point and produces no decision records, the reconciler attributes the kernel observation to the action by the target's identity, which the fixture pins. A Tracee eBPF sensor observes the host kernel throughout the run and is the sole ground truth: decision records are claims, and coverage is computed by reconciling them against kernel observation. The reconciler derives every outcome, checked-allowed, checked-denied, bypassed, prevented, or indeterminate, and no measured component asserts an outcome. The checks cover schema validity, window and accounting consistency, unmatched inspected-class events (which are findings, never noise), id consistency, and denial effectiveness: a checked-denied instance is confirmed by the verified absence of the action's kernel signature, so a denial that did not hold is reported rather than trusted.

Every cell of the coverage grid carries an evidence class. Harness-run cells come from a real harness, and mock-run cells come from the reference path-originator fixture, whose kernel signatures the calibration runs verified against the real harness (Section~\ref{sec:evaluation}). Runs without the kernel sensor take a degraded mode that still executes the full pipeline but labels every output self-reported and unverified.

\subsection{Implementation}
\label{subsec:implementation}

guardscope is a Python framework built around Docker Compose fixtures. The core defines the measurement semantics, the designs, and the reconciler; everything the framework measures or exercises, the harnesses, the controls, and the workloads, is a plugin, and no core code depends on a plugin's identity. A design fixture is a declarative compose base with a control slot, and a control binding is a capability manifest plus the overlays that fill the slot, so a third-party control plugs in without framework changes. A binding whose control emits a proprietary log adds a shim that translates it into the common record schema, with per-record provenance that points back to the vendor log line. Every run pins the harness, the control, the workload, and the sensor images by digest and records them in a run manifest, which makes a measurement reproducible: a third party can re-run the same pins and compare the outcome. guardscope is validated on arm64 (Graviton) and x86\_64 hosts, and the acceptance runs produce identical outcome counts on both (Section~\ref{sec:evaluation}).

\section{Evaluation}
\label{sec:evaluation}

To validate the derived outcomes of Section~\ref{sec:paths} and the guardscope machinery, we run the eight designs as deployments. In this section, we first justify the selection of harnesses and runtime controls, and we then report the settings and the results.

\subsection{Selection of Harnesses and Runtime Controls}
\label{subsec:selection}

\begin{table}[t]
  \centering
  \caption{Tools, roles, demonstrated designs, and versions.}
  \label{tab:tools-and-versions}
  \includegraphics[width=\columnwidth]{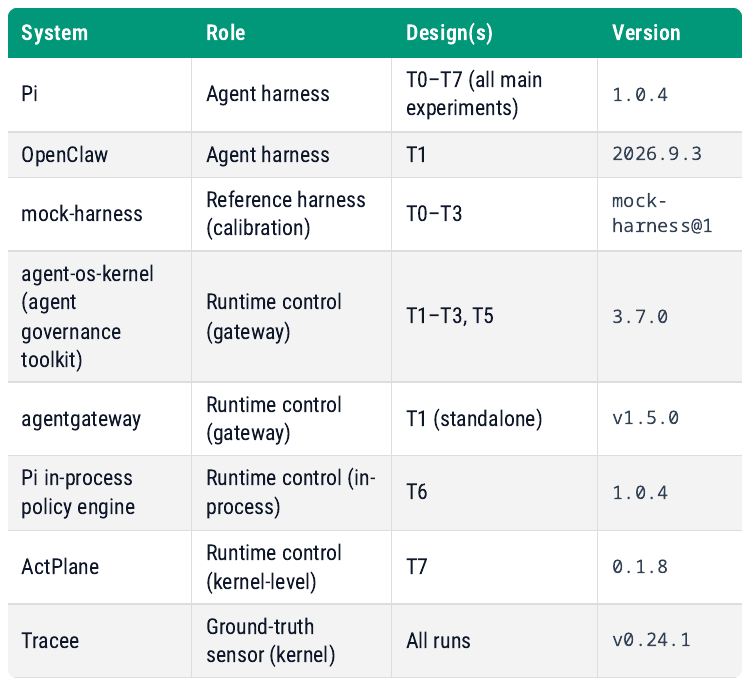}
\end{table}

Table~\ref{tab:tools-and-versions} lists the harnesses, the controls, and the sensor we run, and this subsection states why each was selected.

We select two real harnesses with independent implementations, Pi and OpenClaw, so that a design's measured outcomes can be attributed to its placement rather than to one harness's dispatch implementation. Pi exposes the enforcement hooks that the harness-native and in-process designs require, so every design runs on it. OpenClaw runs the HTTP-gateway design, and its outcome counts match Pi's exactly, which is the harness-independence check. The mock harness is not a harness under measurement; it is the reference path originator that executes the same canonical actions deterministically. Before any mock-derived cell is citable, a calibration gate verifies that its kernel signatures are indistinguishable from the real harness's captures, so the two evidence classes corroborate each other without sharing an implementation.

For the controls, we select maintained, deployed products wherever a design allows an external control, and we cover the eight designs with the fewest products that still exercise both record paths. The gateway designs (T1--T3 and T5) run agent-os-kernel, the enforcement module of Microsoft's agent governance toolkit, a production gateway control. The agentgateway binding adds a second, independent gateway implementation at T1, and its vendor log is translated by a shim rather than emitted natively, so the ingested record path is exercised alongside the native one. The harness-native (T4) and in-process (T6) designs are embedded in the harness by definition, so they use the harness's own mechanisms, the Pi permission chain and the Pi in-process policy engine. The kernel-level design (T7) requires a control that enforces in the kernel, and ActPlane is the only kernel-level system in our classification (Table~\ref{tab:tiermap}), so we bind it.

For the ground truth, the runner manages a single Tracee eBPF sensor per run. The sensor is not a plugin, no measured component can select or modify it, and its health report is a required input of every run.

\subsection{Settings}
\label{subsec:settings}

We run the designs on AWS EC2 Linux hosts, each design brought up as a Docker Compose fixture. The arm64 campaigns run on \texttt{m7g.2xlarge} instances (Graviton3, 8 vCPUs, 32GiB memory, Ubuntu 24.04 LTS, kernel 7.0.x-aws) in the ap-northeast-1 and ap-southeast-1 regions. The x86\_64 validation runs use \texttt{m7i.2xlarge} instances with the same OS on kernel 7.0.0-1015-aws. The kernel-level design (T7) runs on an x86\_64 \texttt{c6i.xlarge} instance with the kernel pinned to 6.17.0-1010-aws, on which ActPlane 0.1.8 loads its eBPF programs in tracepoint mode.\footnote{ActPlane's programs failed to load on the other kernels we tried (7.0.0-1014-aws and 6.8.0-146-generic); we therefore pin the T7 host to 6.17.0-1010-aws.}

\subsection{Results}
\label{subsec:results}

To validate the derived outcomes, we realize each topology design as a guardscope fixture and bind a real runtime control to it. Table~\ref{tab:tools-and-versions} lists the tools, their roles, and the versions we used. The gateway control for T1--T3 and T5 is the binding for agent-os-kernel, the enforcement module of Microsoft's agent governance toolkit; the agentgateway binding, the CNCF gateway project, is demonstrated separately at T1; T4 uses the Pi harness's own tool-call permission chain; T6 configures the harness's in-process policy engine; and T7 binds ActPlane, the kernel-level control. A Tracee sensor records kernel-level events for every run, and we use the record as the ground truth to classify every executed action instance as checked, bypassed, or prevented. Each design runs as a shakedown run followed by an acceptance run: the shakedown run absorbs the first-run effects, such as the harness startup writes that fail the leftover check by design, and only the acceptance runs are counted. We run every design with the Pi harness, the T1 design additionally with the OpenClaw harness, and the T0--T3 designs additionally with the mock harness, whose kernel signatures the calibration runs verified against the real harness. The enforcement-point record comes from the component under test: the gateway decision log, the harness dispatcher hooks, or the kernel-level enforcement log.

Our runs confirmed these derived outcomes for the exercised action classes under every design, including T7, and we see three findings in the run results. First, the mediation boundary of each design matches the derived outcomes for the exercised classes. The T1 runs record decisions for the HTTP tool-call class alone, and the T2 runs additionally prevent the two direct-network actions with the forced-egress configuration; this prevention is kernel-confirmed, so the runs demonstrate the prevented outcome, not only the checked and bypassed ones. The T3 runs add the stdio tool-call class, and the T4--T6 runs mediate both tool-call classes, while the file-write, process-spawn, direct-network, and credential-access classes bypass their enforcement points in every run. In addition, one action in the T6 runs is checked and denied by the in-process policy, which shows that the runs exercise policy decisions, not only allow paths. The T7 run realizes the kernel-level design with ActPlane: six actions are checked and allowed, one is checked and denied, and eight are bypassed. The denied action is a subprocess spawn that ActPlane kills at execution, so the T7 runs also exercise a denial. Two results bound what the T7 run establishes. First, the bypassed set includes all seven tool-call actions, so the kernel position checks a channel only if a deployed hook covers it: the measured per-hook coverage stays below the position bound derived in Section~\ref{sec:paths}. Second, one direct-network action (a UDP send) bypasses ActPlane because of a documented hook gap, which is a property of the evaluated ActPlane hooks rather than of the design.

Second, the checks of a design cover only the channels that cross its enforcement point, and the actions that originate outside those channels take effect without any policy evaluation. In the T5 runs, both tool-call classes are fully mediated, yet the eight actions of the direct-network, process-spawn, file-write, and credential-access classes take effect without any policy evaluation in every run.

Third, the outcome counts are stable across the comparisons the runs support: the T1 column is identical on the Pi and OpenClaw harnesses, and the arm64 and x86\_64 hosts produce identical counts for T0--T3 and T5 with both the real and the mock harness.

The run results also carry a compositional implication. The five paths that bypass every design below the kernel (P2, P5, P11, P12, and P15) cross no enforcement point below the kernel, so no combination of the sub-kernel designs (T1--T6) checks them. Composition therefore adds inspectable features, as the T5 runs show, but not coverage. In contrast, the T7 position bound is blind only to P1, which the harness-native designs (T4--T6) check, so T7 and T4--T6 complement each other at the design level; the measured per-hook coverage of the ActPlane binding (Section~\ref{subsec:results}) stays below this bound.

\subsection{Summary of Findings}
\label{subsec:summary-findings}

Our study yields three key findings. First, coverage follows from where the design puts its enforcement point: a runtime control checks only the paths that cross it, and five of the 15 paths bypass every design below the kernel (Table~\ref{tab:pathmatrix}). The runs confirmed the derived outcomes for the exercised action classes, including a measured T7 run whose per-hook coverage stays below the kernel-position bound, and they showed that the checks cover only the channels that cross the enforcement point. Second, coverage, inspectable features, and exposure to noise are separate dimensions: the design that observes the most paths inspects the least semantic information, and composition adds features without expanding coverage (Table~\ref{tab:topologies}). Third, the deployed solutions concentrate on the designs with narrower coverage (Table~\ref{tab:tiermap}).

These findings indicate that no single control provides both broad path coverage and agent-semantic inspection. Therefore, better runtime control solutions should consider the paths that agent actions could take, and combine tools whose enforcement points sit at different layers, such as harness-native enforcement with kernel-level monitoring.

\section{Limitations and Open Problems}
\label{sec:limitations}

We have identified three limitations: (1) insufficient validation, (2) lack of security effectiveness evaluation, and (3) limits of the kernel-level ground truth.

\subsection{Insufficient Validation}
\label{subsec:validation-boundary}

The empirical validation of Table~\ref{tab:pathmatrix} covers the action classes the reference workload exercises, and each cell of the matrix reflects either a measured acceptance run or a derivation from the design analysis. The workload realizes four of the 15 paths (P2, P4, P6, and P7), so the remaining paths, including every conditional cell, rest on the design analysis alone. The T7 run bounds coverage by the deployed hooks rather than by the kernel position: the measured ActPlane coverage excludes the tool-call classes, and one direct-network action bypasses a documented hook gap.

\subsection{Lack of Security Effectiveness Evaluation}
\label{subsec:coverage-effectiveness}

The main focus of this research was on coverage and security guarantees. ``Checked'' means only that the action's channel crosses an enforcement point, and it does not imply that the control makes a correct decision or that the policy is well configured. ``Prevented'' is likewise an attribute of a blocking mechanism placed outside the policy, such as the forced-egress deny of T2 and T3, rather than of the policy itself. The kernel-level design, in its observation-only form, audits actions but does not stop them; the ActPlane binding we measure enforces, and one action in the T7 run is checked and denied, but a single denial does not establish enforcement effectiveness. Measuring detection accuracy, policy quality, and attack prevention against adversarial workloads is therefore a potential next step.

\subsection{Limits of the Kernel-Level Ground Truth}
\label{subsec:kernel-gt-limits}

Our kernel-level ground truth has its own limits. An eBPF sensor observes kernel-mediated operations, but it does not identify the tool or invocation an operation belongs to by itself, meaning that attributing a syscall record to a specific agent action requires matching against the enforcement-point record. The eBPF sensor also cannot distinguish agent processes from unrelated processes on the host, which makes its record noisy, and it carries no payload content, so an operation's semantics must be inferred from paths, endpoints, and process history. 

\section{Conclusion}
\label{sec:conclusion}
This paper studied the coverage of runtime controls in LLM agent systems: which of an agent's actions a deployed control actually evaluates. We enumerated 15 action paths that span the routes a model's decision can travel from the harness runtime to its effect, identified eight topology designs that span the enforcement positions used in practice, and derived the outcome of each pair of path and design. We then presented guardscope, which turns the analysis into a measurement: the static stage predicts a control's checkable paths from its source, and the empirical stage runs a fixed workload on a deployment that realizes one design and verifies the control's decision records against kernel-level ground truth. The evaluation realizes the eight designs with two agent harnesses and real runtime controls, including Microsoft's agent governance toolkit, the open-source agentgateway project, and ActPlane. The runs confirmed the derived outcomes for the exercised action classes: the checks of a design cover only the channels that cross its enforcement point, five of the 15 paths bypass every design below the kernel, and the measured coverage of the kernel-level control stays below the position bound because it depends on the deployed hooks. The classification of deployed systems shows that most of them concentrate on the designs with the narrowest coverage.

The study has three limitations. The reference workload realizes four of the 15 paths, so the remaining paths rest on the design analysis alone. guardscope measures coverage, not security effectiveness: a checked action may still be allowed by a misconfigured policy, and a single verified denial does not establish enforcement quality. The kernel-level ground truth attributes an operation to an action by target identity or process tree and carries no payload content (Section~\ref{subsec:kernel-gt-limits}). Future work will exercise the conditional paths with dedicated workload instances, evaluate policy effectiveness against adversarial workloads, and extend the attribution of kernel events to agent actions.
\section{AI Usage for This Study}
\label{subsec:ai-usage}
We deploy multiple AI models to assist our study.\footnote{Apart from research assistance, we have used an agent (Pi v1.0.4 with Z.ai's GLM-5.3-flash) to correct grammatical errors and convert tables and figures in LaTeX/tikz to HTML + CSS ones for better readability and visualization.} We use the Pi agent harness (v1.0.4) with Z.ai's GLM-5.3-flash via OpenRouter as a coordinator that interacts with us (human researchers), gets three research agents to work on the investigation tasks and discuss each one's results, and gets us to verify them. For the research agents, we also use Pi as a harness and three models from different vendors: OpenAI's ChatGPT 6.1-Sol, Moonshot AI's Kimi 3, Z.ai's GLM-5.3 (with \texttt{high} reasoning efforts), via AWS Bedrock. The research tasks include (1) the document and source code analysis per project, (2) verification and suggestion on our analysis results (such as potential paths that agents could use, topology designs, classification), and (3) building the testbed environment and the test runs. We, human researchers, check the investigation and compute results.

\bibliographystyle{ACM-Reference-Format}
\bibliography{ref}

\setcounter{section}{0}%
\renewcommand{\thesection}{\Alph{section}}%

\balance

\end{document}